\documentclass[a4paper,15]{article}
\usepackage[utf8]{inputenc}
\usepackage[english]{babel}
\usepackage{csquotes}
\usepackage{array}
\usepackage{url}

\usepackage[a4paper, total={7.25in, 10in}]{geometry}
\usepackage[document]{ragged2e}
\usepackage[sort, numbers]{natbib}
\usepackage{mathtools}
\usepackage{graphicx}
\usepackage{multicol}
\usepackage{listings}
\graphicspath{{/}}
\usepackage{algorithm}
\usepackage{amsmath}
\usepackage{algpseudocode}
\usepackage{fancyhdr}
\usepackage{fixltx2e}

  \date{}
  
  \title{Stats-Calculus Pose Descriptor Feeding A Discrete HMM Low-latency Detection And Recognition System For 3D Skeletal Actions}
\begin{document}

\maketitle

\MakeRobust{\Call}

\makeatletter
\def\BState{\State\hskip-\ALG@thistlm}
\makeatother

\fancypagestyle{firststyle}
{
   \fancyhf{}
   \rhead{Page \thepage}
   \chead{Alexandria, Egypt, Africa}
   \cfoot{\thepage}

}
\thispagestyle{firststyle}
\pagestyle{fancy}
\fancyhf{}
\rhead{Rela-Centro-Stats-Calculus DCT-AMDF}
\lhead{3D skeletal actions detection and recognition}
\cfoot{\thepage}

\providecommand{\keywords}[1]{\textbf{\textit{Keywords---}} #1}

\centering
\large \large \textbf{Rofael Emil Fayez Behnam} \\
Computer and Systems Engineering  \\
Faculty of Engineering \\
Alexandria University- Egypt \\
\url{rofaelemil@gmail.com} 
\justify
    \begin{abstract}
	Recognition of human actions, under low observational latency, is a growing interest topic, nowadays. Many approaches have been represented based on a provided set of 3D Cartesian coordinates system originated at a certain specific point located on a root joint. In this paper, We will present a statistical detection and recognition system using Hidden Markov Model using 7 types of pose descriptors. \\
	\begin{itemize}
		\item Cartesian Calculus Pose descriptor.
		\item Angular Calculus Pose descriptor.
		\item Mixed-mode Stats-Calculus Pose descriptor.
		\item Centro-Stats-Calculus Pose descriptor.
		\item Rela-Centro-Stats-Calculus Pose descriptor.
		\item Rela-Centro-Stats-Calculus DCT Pose descriptor.
		\item Rela-Centro-Stats-Calculus DCT-AMDF Pose descriptor.
	\end{itemize}
	Stats-Calculus is a feature extracting technique, that is developed on Moving Pose descriptor [\citenum{zanfirmoving}], but using a combination of Statistics measures and Calculus measures. \\
	    \end{abstract}

\justify
\section{Introduction}
	Based on a provided set of 3D Cartesian coordinates system [\citenum{zanfirmoving}] originated at a certain specific point located on a root joint, We use Hidden Markov Model to train an efficient skeletal-action recognition and detection system. We will present a new Pose-descriptor set called \textbf{Stats-Calculus} which provides an informative action descriptor for a low latency observational Hidden Markov model system.\\
	Using all 5-consecutive-frame window of Cartesian space measurements to extract this descriptor is an efficient 2-frame lag [\citenum{zanfirmoving}] low-latency observation that can be used to feed a discrete Hidden Markov Model system.	Observations feed to the system is extracted from a continuous real-number space, so a vector quantization algorithms is used as Affinity-Propagation [\citenum{frey07affinitypropagation}] clustering algorithm.\\
	
\section{Related Work}
	Hidden Markov Model is one of stochastic models that is used in this field. (\textit{\citealt{lv2006recognition}}) uses Hidden Markov Models as a weak classifier where its results are supplied to Ada-Boost algorithm. (\textit{\citealt{zanfirmoving}}) presented a Moving Pose descriptor using a non-parametric k-nearest neighbour (kNN) classifier. Many approaches have been presented as using continuous Hidden Markov Model using Gaussian mixture models, Conditional Random Fields and Recurrent Neural Networks.
\section*{3D Skeletal actions Recognition}
\section{Methodology}
\begin{algorithm}[H]
\caption{Stats-Calculus Hidden Markov Model Recoginition}\label{euclid}
\begin{algorithmic}[1]
	\State \Call{Bone Normalization}{}
	\State \Call{Features Extraction}{}
	\State \Call{Dimension Reduction}{}
	\State \Call{Vector Quantization}{}
	\State \Call{Training and Testing Hidden Markov Models}{}
\end{algorithmic}
\end{algorithm}
\section{Bone Normalization}
	In order to avoid variation of bones length due to different actors. So, bones must be scaled to an average learned from data obtained from a sample of different actors. This normalization will affect coordinates of pose, so a topological order must be followed during normalization process. This is same as in  [\citenum{zanfirmoving}]. %%%but it was a defect to specify it as Breadth-First order, which is a special case.
\section{Features Extraction}
	Features vector is constructed using one of six Stats-Calculus descriptors:
	\begin{enumerate}
		\item Cartesian Calculus Pose descriptor.
		\item Angular Calculus Pose descriptor.
		\item Mixed-mode Stats-Calculus Pose descriptor.
		\item Centro-Stats-Calculus Pose descriptor.
		\item Rela-Centro-Stats-Calculus Pose descriptor.
		\item Rela-Centro-Stats-Calculus DCT Pose descriptor.
		\item Rela-Centro-Stats-Calculus DCT-AMDF Pose descriptor.
	\end{enumerate}
\subsection{Cartesian Stats-Calculus Pose descriptor}
	let $ SC(t) $ be Stats-Calculus cartesian descriptor.\\
	let $P(t)$ be pose in Euclidean Cartesian Coordinates (coordinates of joints in a 3D Euclidean space).\\
	let $Stats(P(t))$ be statistical measures of pose [a row vector]. \\
 	let $Calculus(P(t))$ be calculus measures of pose [a row vector].\\
	Then, Stats-Calculus descriptor can be presented as
		$ SC(t) = [ P(t),   Stats(P(t)),  Calculus(P(t)) ] $
	\subsubsection{Calculus Measures}
		Many feature could be extracted from joint motion coordinates. According to Kinematics, velocity and acceleration are two important features of any type of motion [\citenum{zanfirmoving}]. This criteria is only figured in constant-force kinematics system, which is not our case, so we used also Jerd/Jolt "(jerk, jolt (especially in British English), surge or lurch, is the rate of change of acceleration)" which is useful to extract the variation of force applied during an action performing.\\
		\subsubsection*{Numerical Analysis for Calculus Measures}
			Due to using a window of 5 frames [\citenum{zanfirmoving}], It is needed to calculate derivatives through this window.
			let $P(t)$ be pose in Euclidean Cartesian Coordinates.
			Then velocity could be approximated using the highest-order central numerical first derivative over a window of consecutive 5 frames [\citenum{borse1996numerical}].
			$$ V(t) = \frac{P(t_{-2}) - 8 P(t_{-1}) + 8 P(t_{+1}) - P(t_{+2})}{12}$$
			Also, acceleration could be approximated using the highest-order central numerical second derivative over a window of consecutive 5 frames [\citenum{borse1996numerical}].
			$$ A(t) = \frac{-P(t_{-2}) + 16 P(t_{-1}) + 30 P(t) + 16 P(t_{+1}) - P(t_{+2})}{12}$$
			last, jerd/jolt could be approximated using the highest-order central numerical third derivative over a window of consecutive 5 frames [\citenum{borse1996numerical}].
			$$ J(t) = \frac{-P(t_{-2}) + 2 P(t_{-1}) - 2 P(t_{+1}) + P(t_{+2})}{2}$$
			
			then $$  Calculus(P(t)) =   [V(t),  A(t),  J(t)]$$
	\subsubsection{Statistical Measures}
		$P(t)$ is a multi-variate row vector.\\
		Each variable has a real number range, so it could be considered as a random variable, that maps to a hidden probabilistic space with a hidden distribution. \\
		So, Moments could be used as a reasonable measure for $P(t)$ to track variation through out each window of frames.\\
		It is, somehow, "Inference on probabilistic outcome space using sample space".\\
		\begin{itemize}
  			\item Mean ($1^{st}$ central moment - somehow similar to 5 by 1 Gaussian smoothing filter used in [\citenum{zanfirmoving}]).\\
  			It is useful to avoid pose coordinates small variations that happens due to human-action style or nervousness and vibration tensile.
  			\item Variance\\
  			\item Skewness\\
  			 It is useful to detect pose coordinates huge variations.
			\item Kurtosis\\
  			 It is useful to detect movement direction huge rapid variations.
		\end{itemize}
		then $  Stats(P(t)) =   [E(P(t)),  Var(P(t)),  Skew(P(t)),  Kurtosis(P(t))]$
\centering
\onecolumn
\subsection{Angular Stats-Calculus Pose descriptor}
	Same measures, performed on Cartesian Stats-Calculus descriptor, are used on radian angle of spherical coordinates instead of 3d coordinates of joints.\\
	$$ SC(t) = [angles(t), Stats(angles(t)), Calculus(angles(t))]$$ 
	  	\subsubsection{most active bone angles} 
	  	let "Origin" be tha Camera position
	  	as in [\citenum{Sharaf15}].\\
	  	\begin{small}
		\begin{tabular}{l || l | l | l}
			Angle Category & Joint 1 & Vertex & Joint 2\\
	 		\hline
     		\hline
			Right   & Shoulder Center & Shoulder Right & Elbow Right \\
			Arm			 & Shoulder Right & Elbow Right & Wrist Right \\
			Angles		 & Elbow Right  & Wrist Right & Hand Right \\
		     \hline
		     \hline
			Left    &  Shoulder Center  &  Shoulder Left & Elbow Left \\
			Arm		 & Shoulder Left &  Elbow Left &  Wrist Left\\
			Angles	 & Elbow Left &  Wrist Left &  Hand Left \\
		     \hline
		     \hline
			Right   &  Hip Center &  Hip Right &  Knee Right\\
			Leg	 & Hip Right &  Knee Right &  Ankle Right\\
			Angles	 & Knee Right &  Ankle Right &  Foot Right\\
		     \hline
 		     \hline
			Left &  Hip Center &  Hip Left &  Knee Left\\
			Leg	 & Hip Left &  Knee Left &  Ankle Left\\
			Angles	 & Knee Left &  Ankle Left &  Foot Left\\
		     \hline
		     \hline
			Symmetric  &  Shoulder Left &  Shoulder Center &  Shoulder Right\\
			Angles	 & Head &  Shoulder Center &  Spine\\
							 & Shoulder Center &  Spine &  Hip Center\\
		     \hline
		     \hline
			Big   &  Elbow Left &  Shoulder Center & Elbow Right\\
			Symmetric	 & Wrist Left &  Shoulder Center & Wrist Right\\
			Angles & Knee Left &  Hip Center &  Knee Right\\
								 & Ankle Left &  Hip Center &  Ankle Right\\
		     \hline
		     \hline
			Big   & Wrist Left & Hip Center & Ankle Left\\
			Hand-	& Wrist Right & Hip Center & Ankle Right\\
			Foot	&  Wrist Left &  Hip Center &  Ankle Right\\
			Angles	 & Wrist Right &  Hip Center &  Ankle Left\\
		     \hline
		     \hline
					 &  Wrist Left &  Origin&  Shoulder Center \\
					 & Elbow Left&  Origin &  Shoulder Center \\
					 & Shoulder Left &  Origin&  Shoulder Center\\
					& Ankle Left &  Origin&  Hip Center \\
					 & Knee Left &  Origin&  Hip Center\\
			Camera-	 & Hip Left &  Origin&  Hip Center\\
			Facing	 & Wrist Right &  Origin&  Shoulder Center\\
			Angles	 & Elbow Right &  Origin&  Shoulder Center\\
					 &  Shoulder Right &  Origin&  Shoulder Center\\
					 & Ankle Right &  Origin&  Hip Center\\
					 & Knee Right &  Origin &  Hip Center \\
					 & Hip Right &  Origin &  Hip Center \\
		\end{tabular}
	\end{small}
	\\
	\begin{figure}[h]
	\centering
	  \includegraphics[scale=0.45]{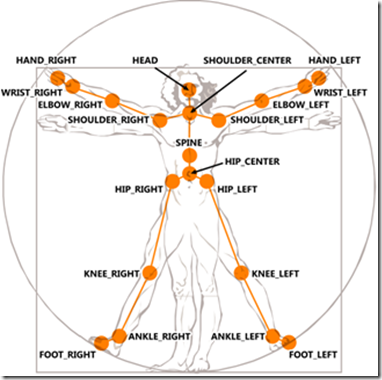}
	\caption{joints}
	\end{figure}
\justify

\subsection{Mixed-mode Stats-Calculus Pose Descriptor}
	It is formed from using similar features from both Cartesian Coordinates Stats-Calculus Pose descriptor and Angular Stats-Calculus Pose Descriptor.
	$$ Mixed(t) = [P(t),angles(P(t))]$$
	$$ SC(t) = [Mixed(t), Stats(Mixed(t)), Calculus(Mixed(t))]$$ 

\subsection{Centro-Stats-Calculus Pose descriptor}
	For some moving actions as running, walking and some games as Fighting, It is needed to track position of the actor and his/her relative position during performing an action. So, no need to select a joint as origin for the system, but it is needed to define a dynamic positioned origin. our claim is using the centroid of all joints through all each action instance to be the origin for the system.
	
	$$ Centro(X,Y,Z) = \frac{\sum_{n=1}^{no. of frames} \sum_{j=1}^{no. of joints} Joint(j,n,X,Y,Z)}{no.of frames * no. of joints}$$ 
	
	Same measures, performed on Cartesian Stats-Calculus descriptor, are used on $ Centro(X,Y,Z)$ instead of $P(t)$.
	$$ SC(t) = [Centro(X,Y,Z), Stats(Centro(X,Y,Z)), Calculus(Centro(X,Y,Z))]$$ 
\subsection{Rela-Centro-Stats-Calculus Pose descriptor}
	For free moving actions, It is recommended to have a position tracker alongside a pose tracker, so a mix of Centro- and Cartesian Stats-Calculus is recommended for this case. \\
		$$ CentroSC(t) = [Centro(X,Y,Z), Stats(Centro(X,Y,Z)), Calculus(Centro(X,Y,Z))]$$ 
		$$ CartesianSC(t) = [ P(t),   Stats(P(t)),  Calculus(P(t)) ] $$
		$$ SC(t) = [CentroSC(t), CartesianSC(t)] $$
\subsection{Rela-Centro-Stats-Calculus DCT Pose descriptor}
	using transforms to uniquely differentiate between different poses is recommended.
	\begin{itemize}
		\item  calculate Discrete Cosine transform (DCT) of \textbf{Rela-Centro-Stats-Calculus} and truncating it to 100 dimension.
	\end{itemize}
		$$ SC(t) = DCT([CentroSC(t), CartesianSC(t)]) $$

\subsection{Rela-Centro-Stats-Calculus DCT-AMDF Pose descriptor}
	using transforms to uniquely differentiate between different poses is recommended.
	\begin{itemize}
		\item First, calculate Discrete Cosine transform of \textbf{Rela-Centro-Stats-Calculus} and truncating it to 100 dimension.
		\item Second, calculating its average magnitude difference function (AMDF) at $ N=45 $ as in [\citenum{apte2012speech}].
	\end{itemize}
	$$ DCTSC(t) = DCT([CentroSC(t), CartesianSC(t)]) $$
	$$ AMDF(t,k) = \frac{\sum_{n=1}^{N}{|DCTSC(t,n)-DCTSC(t,n+k)|}}{N} $$

\section{Dimension Reduction}
	All Stats-Calculus Pose descriptor are normalized, in order to not bias to any component. \\
	Due to high-dimensionality of features vector, so Principal Component Analysis is used to get efficient dimension reduction.

\section{Vector Quantization}
\begin{itemize}
	\item Affinity-Propagation [\citenum{frey07affinitypropagation}] clustering algorithm is used to cluster similar pose descriptors, in order to use clusters identifiers as a representative to all its cluster pose descriptors during training and testing using Hidden Markov Model.
\end{itemize}

\section{Training and testing Hidden Markov Model}
\begin{itemize}

	\item  Hidden Markov Model is trained using sequences of cluster labels of each frame descriptor for each performed action using Expectation-Maximization algorithm (so-called Baum-Welch) to maximize likelihood of sample learned features to get a Hidden Markov Model recognizer for each action. \\
	\item  Observations feed to Hidden Markov Model are sequences of clusters identifiers. 
	\item  Viterbi (Dynamic Programming) algorithm is used to decode Hidden Markov Models system. \\
	\end{itemize}

\justify
\section{Recognition Experiments}
	This technique is tested on different data-sets:
	\begin{itemize}
		\item UTKinect-Action Dataset [\citenum{xia2012view,devanne2013space}]
		\item MSR-Action3D Dataset [\citenum{wang2012mining,HusseinTGE13}]
		\item MSR Daily Activity 3D Dataset [\citenum{wang2012mining}]
		\item Florence 3D actions Dataset [\citenum{seidenari2013recognizing,devanne20143}]
	\end{itemize}
	
	\subsection{UTKinect-Action Dataset [\citenum{xia2012view,devanne2013space}]}
	"The videos was captured using a single stationary Kinect with Kinect for Windows SDK Beta Version. There are 10 action types: walk, sit down, stand up, pick up, carry, throw, push, pull, wave hands, clap hands. There are 10 subjects, Each subject performs each actions twice."
	\begin{figure}[h]
	\centering
	  \includegraphics[scale=0.2]{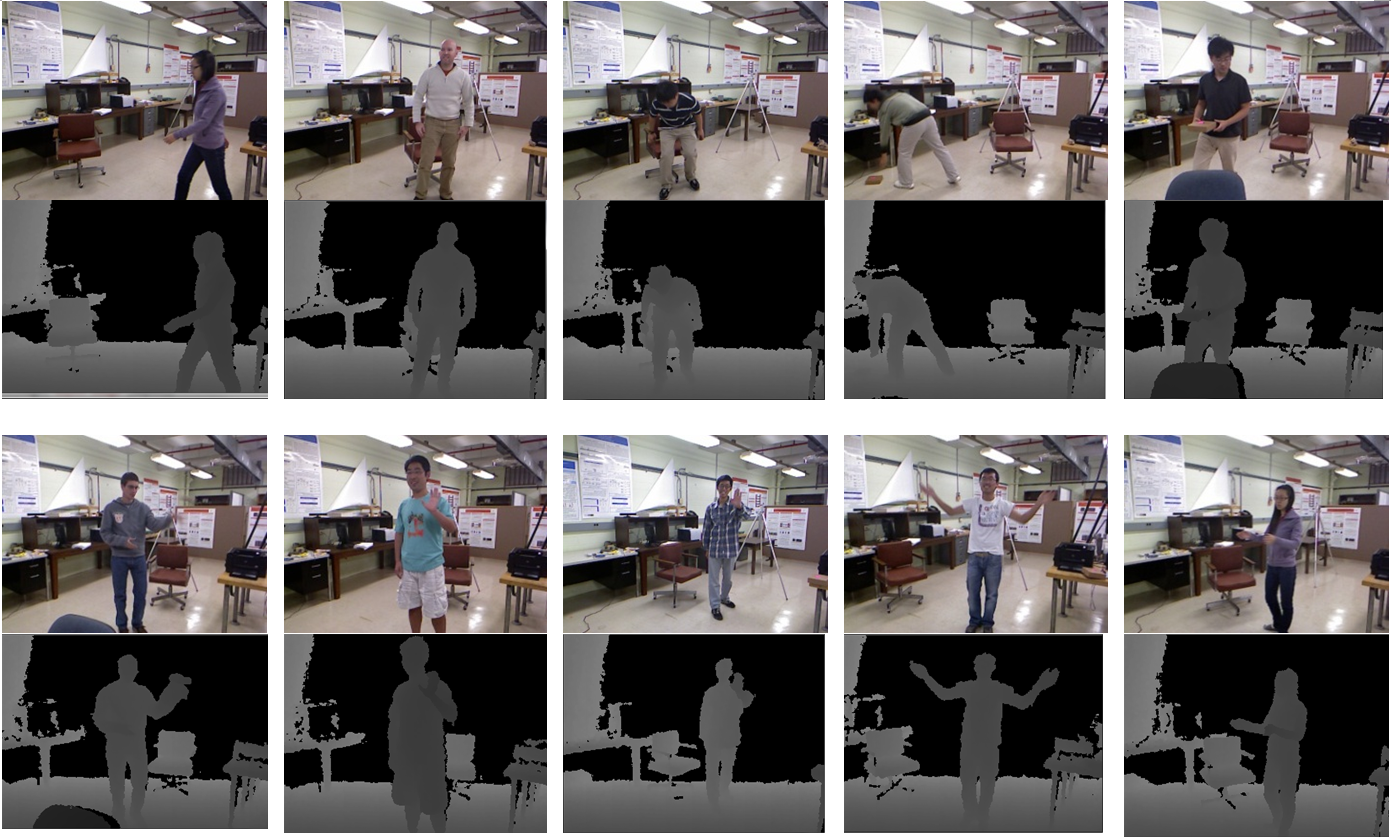}
	\caption{UTKinect-Action Dataset sample}
	\end{figure}
	\subsubsection{3D Skeletal Action Recognition}
	* Experimental Setup: using subjects 1-5 for training and subjects 6-10 for testing.\\
	\begin{tabular}{| l || c |}
	  	\hline
  		descriptor  & accuracy \\
  		\hline
  		\hline
  		HOJ3D [\citenum{xia2012view}] & 90.92\%  \\
  		Space-time Pose [\citenum{devanne2013space}] & 91.5\% \\
  		\hline
  		Centro-Stats-Calculus & 68.8889\% \\
  		Rela-Centro-Stats-Calculus & 80.0000\% \\
  		Rela-Centro-Stats-Calculus DCT-AMDF & 97.7778\% \\
  		\hline
	\end{tabular}
			\subsection{MSR-Action3D Dataset [\citenum{wang2012mining,HusseinTGE13}]}
		It consists of 20 actions performed 2-3 times by each of 10 subjects. \\
		Some files are corrupted, according to [\textit{\citealt{wang2012mining, HusseinTGE13}}]\\
			\textit{a02\_s03\_e02\_skeleton3D, a04\_s03\_e01\_skeleton3D,\\
			a07\_s04\_e01\_skeleton3D, a13\_s09\_e01\_skeleton3D, \\
			a13\_s09\_e02\_skeleton3D, a13\_s09\_e03\_skeleton3D,\\
			a14\_s03\_e01\_skeleton3D, a20\_s07\_e01\_skeleton3D,\\
			a20\_s07\_e03\_skeleton3D and a20\_s10\_e03\_skeleton3D} \\

		\subsubsection{3D Skeletal Action Recognition}
		It is split into 3 overlapping action sets [$AS1,AS2 and AS3$], as in [\citenum{GowayyedTHE13,padilla2014discussion,li2010action,HusseinTGE13,devanne2013space}] \\
		Two experimental setup are used
		\begin{itemize}
		\item subjects 1,2,3,4,5 for training and 6,7,8,9,10 for testing as in [\citenum{oreifej2013hon4d}] \\
			\begin{tabular}{| l || c |}
	  			\hline
  				descriptor  & Coordinates Stats-Calculus \\
  				\hline
  				\hline
  				AS1 &  87.619\% \\
  				AS2 &  86.61\% \\
  				AS3 &  94.59\% \\
  				\hline
  				\hline
  			overall &  89.6\%\\
  			  	\hline
			\end{tabular}
			\item subjects 1,3,5,7,9 are used for training and 2,4,6,8,10 are used for testing as in [\citenum{wang2012mining,li2010action}] \\
			\begin{tabular}{| l || c | r |}
	  			\hline
  				descriptor  & Coordinates Stats-Calculus & Centro-Stats-Calculus Pose \\
  				\hline
  				\hline
  				AS1 &  83.8095\% & 81.9048\% \\
  				AS2 &  86.6071\% & 85.7143\% \\
  				AS3 &  92.7928\% & 85.5856\% \\
  				\hline
  				\hline
  			overall &  87.7365\% & 84.4015\% \\
  			  	\hline
			\end{tabular}
			\begin{figure}[h]
	\centering
	 \includegraphics[scale=0.5]{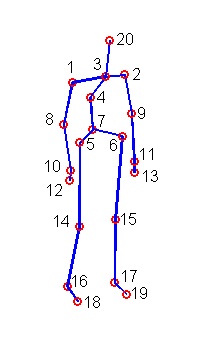}
	\caption{MSR-Action3D Dataset sample}
	\end{figure}
		\end{itemize}
		\begin{tabular}{| l | c |}
	  	\hline
  		method  & accuracy \\
  		\hline
  		\hline
  		Recurrent Neural Network [\citenum{wang2012mining,martens2011learning,zanfirmoving}] & 42.5\% \\
  		Dynamic Temporal Warping [\citenum{wang2012mining,muller2006motion,zanfirmoving}] & 54.0\% \\
  		Hidden Markov Model [\citenum{lv2006recognition,zanfirmoving,wang2012mining}] & 63.0\% \\
  		Latent-Dynamic CRF [\citenum{morency2007latent,zanfirmoving}] & 64.8\% \\
  		Latent-Dynamic Canonical Poses [\citenum{zanfirmoving}] & 65.7\% \\
  		Action Graph on Bag of 3D Points [\citenum{wang2012mining,zanfirmoving}] & 74.7\% \\
  		EigenJoints [\citenum{zanfirmoving}] & 81.4\% \\
  		Actionlet Ensemble [\citenum{wang2012mining}] & 88.2\% \\
  		MP [\citenum{zanfirmoving}] & 91.7\% \\
  	  	\hline
  	  	Coordinates Stats-Calculus &  89.6\%  \\
  		\hline
  	\end{tabular}
  	\\
	\subsection{MSR Daily Activity 3D Dataset [\citenum{wang2012mining}]}
	"The MSR Daily Activity 3D dataset is a most widely applied benchmark dataset in human behavior understanding tasks. This dataset contains 16 human activities, performed by 10 subjects. Each subject performs each activity twice, once in standing position and once in sitting position."
	\subsubsection{3D Skeletal Action Recognition}

	\begin{tabular}{| l | c | }

  		\hline
  		 method & accuracy   \\
  		\hline
  		\hline
  		Dynamic Temporal Warping [\citenum{muller2006motion,zanfirmoving}] & 54.0\% \\
  		Actionlet Ensemble (3D pose only) [\citenum{wang2012mining,zanfirmoving}] &68.0\% \\
  		MP [\citenum{zanfirmoving}] &73.8\% \\
  		\hline
  		Rela-Centro-Stats-Calculus DCT-AMDF &  87.5\%  \\
  		\hline
	\end{tabular}
	
	\subsection{Florence 3D actions Dataset [\citenum{seidenari2013recognizing,devanne20143}]}
		"The dataset collected at the University of Florence during 2012, has been captured using a Kinect camera. It includes 9 activities: wave, drink from a bottle, answer phone,clap, tight lace, sit down, stand up, read watch, bow. During acquisition, 10 subjects were asked to perform the above actions for 2/3 times. This resulted in a total of 215 activity samples."\\
	* Experimental Setup: using 17 instance of each action for training and the remaining for testing
	\subsubsection{3D Skeletal Action Recognition}
	\begin{tabular}{| l || c |}
	  	\hline
  		descriptor  & accuracy \\
  		\hline
  		Bag-of-poses[\citenum{seidenari2013recognizing}] & 82\% \\
  		\hline
  		Rela-Centro-Stats-Calculus DCT-AMDF & 58.065\% \\
  		Rela-Centro-Stats-Calculus DCT & 68.8889\% \\
  		Rela-Centro-Stats-Calculus & 74.194\% \\
		Angular-Stats-Calculus & 74.194\% \\

  		\hline
	\end{tabular}

	%%%%%%%%%%%%%%%%%%%%%%%%%%%%%%%%%%%%%%%%%%%%%%%%%%
	%%%%%%%%%%%%%%%%%%%%%%%%%%%%%%%%%%%%%%%%%%%%%%%%%%
	%%%%%%%%%%%%%%%%%%%%%%%%%%%%%%%%%%%%%%%%%%%%%%%%%%
	\section*{3D Skeletal actions Detection using unsegemented sequences}
		adapted from the idea of segementation$-$free OCR [\citenum{plotz2009markov}]
	\section{Methodology}
		\begin{algorithm}[H]
		\caption{Stats-Calculus HMM Detection}\label{euclid}
			\begin{algorithmic}[1]
				\State \Call{Bone Normalization}{}
				\State \Call{Features Extraction}{}
				\State \Call{Dimension Reduction}{}
				\State \Call{Vector Quantization}{}
				\State \Call{Training HMMs}{}
				\State \Call{Parallel Connections of HMMs}{}
			\end{algorithmic}
		\end{algorithm}

\subsection{Dimension Reduction}
	All Stats-Calculus Pose descriptor are normalized, in order to not bias to any component. \\
	Due to high-dimensionality of features vector, so Principal Component Analysis is used to get efficient dimension reduction.

\subsection{Vector Quantization}
\begin{itemize}
	\item Affinity-Propagation [\citenum{frey07affinitypropagation}] clustering algorithm is used to cluster similar pose descriptors, in order to use clusters identifiers as a representative to all its cluster pose descriptors during training and testing using Hidden Markov Model.
\end{itemize}

\subsection{Training and testing Hidden Markov Model (HMM)}
\begin{itemize}

	\item  Hidden Markov Model is trained using sequences of cluster labels of each frame descriptor for each performed action using Expectation-Maximization algorithm (so-called Baum-Welch) to maximize likelihood of sample learned features to get a Hidden Markov Model recognizer for each action. 
	\item  Observations feed to Hidden Markov Model are sequences of clusters identifiers. 
	\item  Viterbi (Dynamic Programming) algorithm is used to decode Hidden Markov Models system. 
	\end{itemize}

	\subsection{Parallel Connections of Hidden Markov Models (HMM)} 
	as in segmentation-free OCR [\citenum{plotz2009markov}]  where each action previously trained Hidden Markov Model is a unit in the structure of a bigger Hidden Markov Model as showed in Figure 4\\
	\begin{figure}[h]
	\centering
	\includegraphics[scale=0.65]{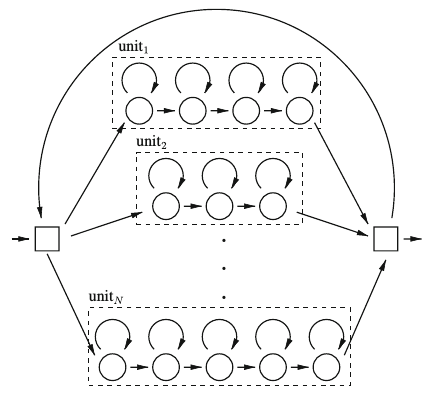}

	\caption{Parallel Connections of Hidden Markov Models (HMM)}
	\end{figure}

\section{Detection Experiments using unsegemented sequences}
	This technique is tested on different data-sets:
	\begin{itemize}
		\item UTKinect-Action Dataset [\citenum{xia2012view,devanne2013space}]
		\item MSR-Action3D Dataset [\citenum{wang2012mining,HusseinTGE13}]
		\item Florence 3D actions Dataset [\citenum{seidenari2013recognizing,devanne20143}]
	\end{itemize}
	\subsection{UTKinect-Action Dataset [\citenum{xia2012view,devanne2013space}]}
	* Experimental Setup: using subjects 1-3 for training and subjects 4-10 for testing.\\
	* using a sliding-window of width = 7 frames, an Hidden MArkov Model (HMM) system of 3 states for each action. using Angular Stats-Calculus Pose Descriptor. \\
	* Results: $$ precision = 0.9984, recall = 0.9753, F_1-Score = 0.9867 $$
	\subsection{MSR-Action3D Dataset}
	* Experimental Setup: using subjects 1-3 for training and subjects 4-10 for testing.\\
	* using a sliding-window of width = 3 frames, an Hidden MArkov Model (HMM) system of 3 states for each action. using Angular Stats-Calculus Pose Descriptor. \\
	* Results: 
	\begin{itemize}
		   \item AS1: $$ precision = 1.0, recall = 1.0, F_1-Score = 1.0 $$
		   \item AS2: $$ precision = 1.0, recall = 0.9996, F_1-Score = 0.9998 $$
		   \item AS3:  $$ precision = 1.0, recall = 0.9996, F_1-Score = 0.9998 $$
	\end{itemize}
	\subsection{Florence 3D actions Dataset [\citenum{seidenari2013recognizing,devanne20143}]}
	* Experimental Setup: using instances 1-4 for training and instances 5-18 for testing.\\
	* using a sliding-window of width = 3 frames, an Hidden MArkov Model (HMM) system of 3 states for each action. using Stats-Calculus Pose Descriptor. \\
	* Results: $$ precision = 1.0, recall = 0.4210, F_1-Score = 0.5926 $$

\section{Conclusion and Comparison}
	* A novel of Stats-Calculus Moving Pose descriptors framework is presented for both 3D skeletal action detection and recognition which uses pose, calculus kinematic information and statistical measures. 
	
	\subsection{3D Skeleton actions Detection}
	\begin{tabular}{| l || c | r |}
	  	\hline
  		method  & Stats-Calculus & MP [\citenum{zanfirmoving}] \\
  		\hline
  		\hline
  		MSR-Action 3D [\citenum{wang2012mining,HusseinTGE13}] &  0.9998 & 0.892 \\
  		\hline
  	\end{tabular}

\section{Future Enhancement}
	* adding 2 extra angles for efficient Head-gestures as nodding
	\begin{itemize}
		\item angle of: head---Shoulder Center---Shoulder Left.
		\item angle of: head---Shoulder Center---Shoulder Right. 
	\end{itemize}
\bibliographystyle{plainnat}
\bibliography{statscalculus}
\begin{minipage}{0.5\textwidth}
\large \large Author \\
\large \large \textbf{Rofael Emil Fayez Behnam} \\
Computer and Systems Engineering  \\
Faculty of Engineering \\
Alexandria University- Egypt \\
\url{rofaelemil@gmail.com} \\
\textit{\textbf{ Bachelor of Computer and Systems Engineering,}} $4^{th}$ year, Computer and Systems Engineering department, Faculty of Engineering, Alexandria University [2011- 2016]\\
expected July 2016 
\textbf{(86.8\%$-$ Distinction)}
\end{minipage} \hfill
\begin{minipage}{0.45\textwidth}
    \includegraphics[width=0.45\textwidth]{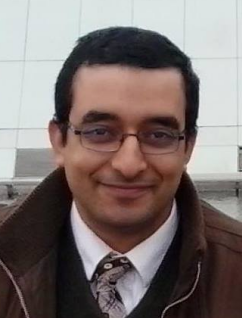}
\end{minipage}

\end{document}